\documentclass{article}
\usepackage{ijcai22}

\usepackage{times}
\usepackage{soul}
\usepackage{url}
\usepackage[hidelinks]{hyperref}
\usepackage[utf8]{inputenc}
\usepackage[small]{caption}
\usepackage{graphicx}
\usepackage{amsmath}
\usepackage{amsthm}
\usepackage{booktabs}
\usepackage{algorithm}
\usepackage{algorithmic}
\usepackage{amssymb}
\usepackage{booktabs}
\usepackage{algorithm}
\usepackage{algorithmic}
\usepackage{float}
\usepackage{subfig}
\newtheorem{definition}{Definition}
\usepackage{xcolor}
\usepackage{todonotes}
\usepackage{color}

\title{Multi-Graph Fusion Networks for Urban Region Embedding}

\author{
Shangbin Wu\footnote{These authors contributed equally to this work}$^1$\and
Xu Yan\footnotemark[1]$^1$\and
Xiaoliang Fan\thanks{Corresponding author: fanxiaoliang@xmu.edu.cn}$^{1}$\and
Shirui Pan$^2$\and
Shichao Zhu$^3$\and \\
Chuanpan Zheng$^1$\and
Ming Cheng$^1$\and
Cheng Wang$^1$
\affiliations
$^1$Fujian Key Laboratory of Sensing and Computing for Smart Cities, School of Informatics, Department of Computer Science and Technology, Xiamen University, China\\
$^2$Department of Data Science and AI, Faculty of Information Technology, Monash University, Australia \\
$^3$University of Chinese Academy of Sciences\\
\emails
\{shangbin, yanxu97, zhengchuanpan\}@stu.xmu.edu.cn,
\{fanxiaoliang, chm99, cwang\}@xmu.edu.cn
shirui.pan@monash.edu,
zhushichao@iie.ac.cn
}

\begin{document}
\maketitle

\begin{abstract}
Learning the embeddings for urban regions from human mobility data can reveal the functionality of regions, and then enables the correlated but distinct tasks such as crime prediction. Human mobility data contains rich but abundant information, which yields to the comprehensive region embeddings for cross domain tasks. In this paper, we propose multi-graph fusion networks (MGFN) to enable the cross domain prediction tasks. First, we integrate the graphs with spatio-temporal similarity as mobility patterns through a \emph{mobility graph fusion module}. Then, in the \emph{mobility pattern joint learning module}, we design the multi-level cross-attention mechanism to learn the comprehensive embeddings from multiple mobility patterns based on intra-pattern and inter-pattern messages. Finally, we conduct extensive experiments on real-world urban datasets. Experimental results demonstrate that the proposed MGFN outperforms the state-of-the-art methods by up to 13.11$\%$ improvement. https://github.com/wushangbin/MGFN





\end{abstract}

\section{Introduction} \label{sec.1}
  
Revealing urban region embedding aims to learn quantitative representations of regions from multi-sourced data, such as Point-of-Interests (POI), check-in, and human mobility~\cite{HDGE2017}. Human mobility data reflects the human interactions and cooperation, and thus can be used to conduct distinct tasks such as epidemic~\cite{Wu2020}, economics~\cite{multigraph_economy}, crime~\cite{crime-ST-SHN} prediction, etc.

The cross-domain downstream tasks such as crime prediction and check-in prediction, are used to verify the effectiveness of region embeddings. Existing studies ~\cite{beyondGeo,HDGE2017,ZE-Mob2018} taking all detailed time-series mobility records as input, could merely learn a specific representation (e.g., change of traffic flows), rather than generalized urban region embeddings. 

Human mobility data contains both abundant information and complex patterns~\cite{new_mobility}, which yields to the comprehensive region embeddings for cross domain tasks. For example, Figure~\ref{fig.1} (left) shows two repeated patterns, which can be integrated as Pattern 1 (morning peak) and Pattern 2 (weekend) respectively. Figure~\ref{fig.1} (right) shows the complexity of mobility patterns that there are two distinct patterns (i.e., variation and similarity). 
Measuring a generalized region embeddings from abundant human mobility data is challenging:


 
\begin{figure}[t]
	\centerline{\includegraphics[width=0.5\textwidth]{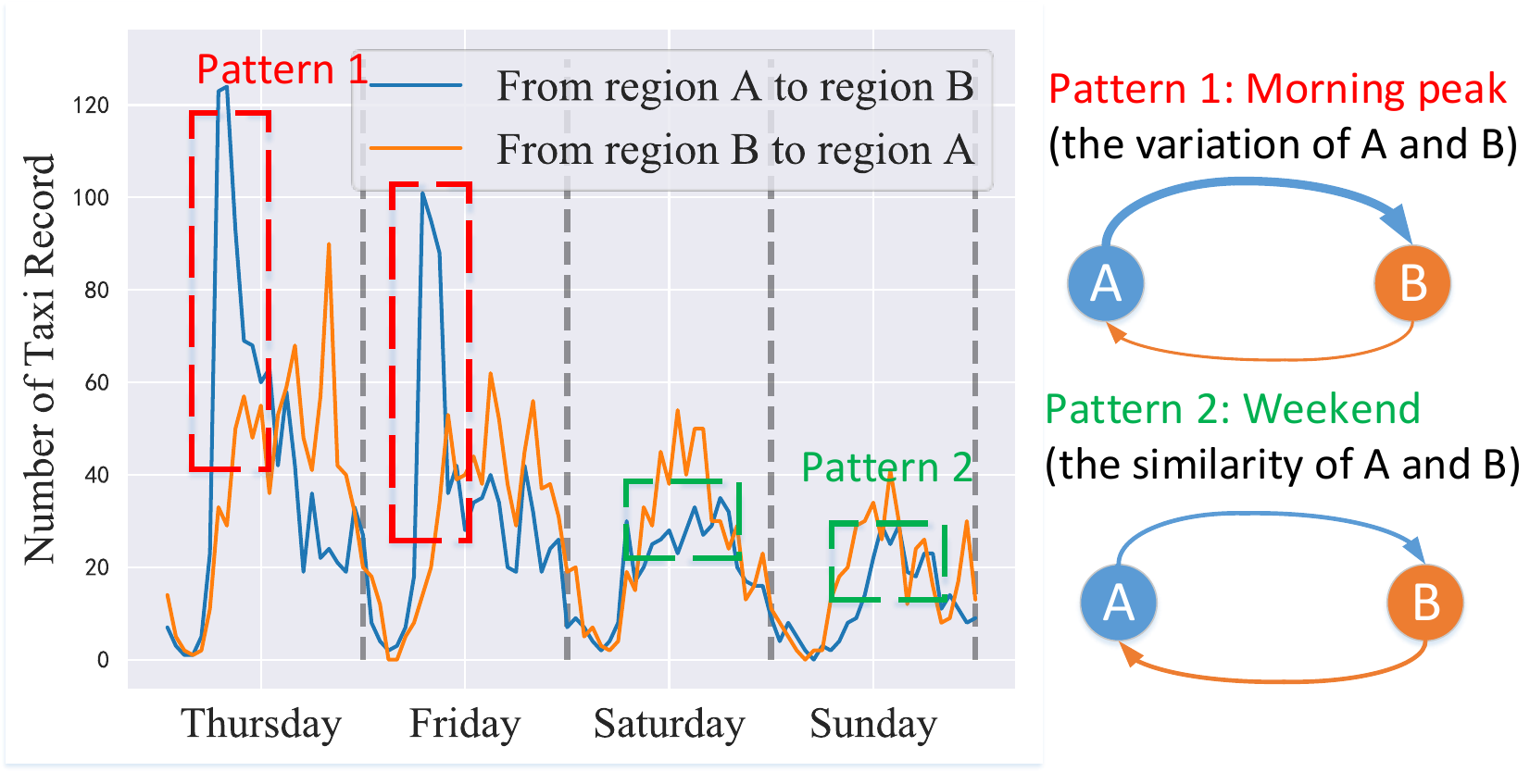}}
	\caption{A motivating example. The complexity of mobility patterns depends on both urban periodicity and regional functionality distribution. For two regions (A is a residential area, and B is a office area), there could be two distinct patterns (i.e., variation, similarity).}
	\label{fig.1}
\end{figure} 

\textbf{Challenge 1: }How to process the fine-grained mobility data to learn a generalized embedding? In our considered scenario, learning from abundant human mobility data may only attain the change of mobility flows, rather than a generalized embedding. 

\textbf{Solution 1: Mobility graph fusion.} To attain an effective and generalized embedding from abundant human mobility data, we first approach regions as interactive and interdependent nodes by constructing the human mobility data as mobility multi-graph. Then, we aggregate mobility graphs as selected mobility patterns according to the spatio-temporal distance between any two mobility graphs. 

\textbf{Challenge 2: }How to jointly learn from mobility patterns? Training a separate pattern mining model on each of them may not capture comprehensive representation of regional characteristics. 

\textbf{Solution 2: Mobility Pattern Joint Learning.} The constructed mobility graphs possess regularized characteristics, which are fully connected, directed, and weighted with multiple edges (i.e., multi-graph). Different from previous graph representation models~\cite{MVURE2020}, we take advantage of above characteristics by designing two modules: (1) Intra-pattern message passing, which utilizes structural information inside each graph to learn a local embedding; and (2) Inter-pattern message cross attention, which conducts attention mechanism among different graphs to jointly learn comprehensive region embeddings.

The major contributions of this paper are:
\begin{itemize}
	\item We study the urban region representation problem on fine-grained human mobility data, and propose a mobility graph fusion module with spatio-temporal dependencies where redundant graphs are integrated as patterns.
	\item We propose a mobility pattern joint learning module, which learns the region embedding from intra-pattern message and inter-pattern message simultaneously in a new manner, 
	with the hope that the cross-graph information can mutually enhance each other.  
	\item Extensive experimental results show that our mobility graph fusion method can effectively uncover the complex mobility patterns. And our method outperforms state-of-the-art baselines up to 13.11$\%$ in crime and check-in prediction tasks in terms of various metrics.  
\end{itemize}



\section{Problem Statement} \label{problem_statement}

We provide necessary preliminary concepts in this work, and formalize the problem of urban region embedding.


\begin{definition}[Mobility Graph]
The mobility graph at the time step $t$ is defined as a directed and weighted graph $G_t = (V, E_t)$, where $V$ denotes the node set with node $v_i \in V$ representing region $v_i$, and $E_t$ denotes the edge set with edge $e^t_{ij} = (v_i, v_j, \omega^{t}_{ij}) \in E_t$ representing the number of people $\omega_{ij}$ move from urban region $v_i$ to $v_j$ at time $t$. 
\end{definition}

\begin{definition}[Mobility Multi-graph]
It is defined as a directed and weighted multi-graph $\mathrm{G} = \cup_{t=0}^{T-1} \left \{ G_t = (V, E_t) \right \} $, where $G_t$ is a mobility graph at time $t$, and $V$ denotes the node set that corresponds to regions, and $E_t$ denotes the edge set that corresponds to mobility condition at time $t$. 
\end{definition}

\begin{definition}[Mobility Pattern]
The mobility patterns $\mathcal{G} = \{ \mathcal{G}_0, \mathcal{G}_1, ..., \mathcal{G}_{N-1} \}$ are the result of fusing similar mobility multi-graph. A pattern $\mathcal{G}_k$ is also a directed and weighted graph with the same node set $V$ as mobility multi-graph. 

\end{definition}

\begin{definition}[Urban Region Embedding]
Given mobility multi-graph $\mathrm{G}$, the goal of urban region embedding is to learn a mapping function $\phi: v_i \rightarrow \mathbb{R}^{d}$ to generate low dimensional embedding $\hat{H} \in \mathbb{R}^{ |V| \times d}$ of each region $v_i \in V$, where $d$ represents the dimension of urban region embeddings.
\end{definition}

\section{Methodology}
\begin{figure}[htbp]
	\centerline{\includegraphics[width=0.5\textwidth]{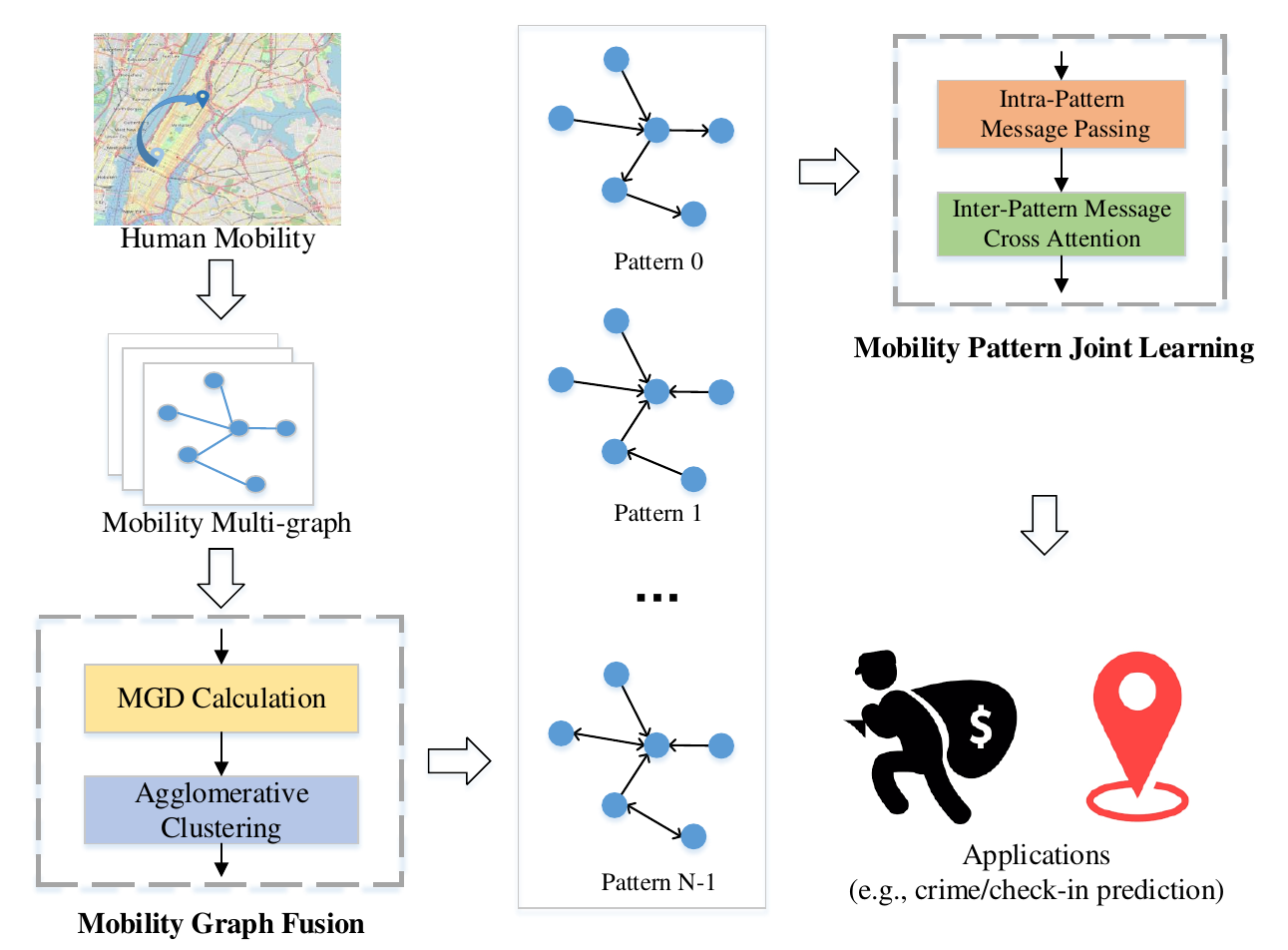}}
	\caption{MGFN Framework. The architecture transforms human mobility patterns into representation abilities for downstream tasks (e.g., crime, check-in prediction). Our framework consists of two modules: 1) Mobility Graph Fusion module where time-series multi-graph are fused by a Mobility Graph Distance (MGD) measurement method; and 2) Mobility Pattern Joint Learning module (detailed in Figure~\ref{fig.3}) which learns the region embedding by both intra-pattern message and inter-pattern message. }
	\label{fig.2}
\end{figure}

In this section, we first introduce the mobility graph fusion module, in which we propose a novel mobility graph distance to measure the similarity between different mobility graphs. Then, we present an effective mobility pattern joint learning module, which contains intra-pattern message passing and inter-pattern message cross attention to capture comprehensive regional characterise by mobility patterns. Figure~\ref{fig.2} shows the overall architecture of our multi-graph fusion network. 

\subsection{Mobility Graph Fusion}\label{sec.3.1}
Human mobility reveals the functions and properties of urban regions~\cite{HDGE2017}. Our intuition is that mobility patterns are able to describe the urban region functionality thus in favour of learning the generalized representation. For example, estimate whether a region is work area or residential area according to the human mobility direction during the morning peak hours (i.e., 7-9 a.m.). Therefore, instead of learning how human mobility changes, we model the problem as time-series mobility multi-graph fusion (MGF) to extract the mobility patterns. First, we define mobility graph distance (MGD) to calculate the distance between mobility graphs, and apply clustering methods with MGD to cluster similar mobility graphs. Second, in each cluster, we aggregate edges in all mobility graphs to form a mobility pattern.
\subsubsection{Spatial Structure Distance on Mobility Graph}
Data with similar mean and variance may have higher similarity. Specifically, for a mobility graph $G$, we assume that the weight $w$ of edge $e = (u, v, w)$ is sampled from a Gaussian distribution, and calculate its mean $ \mu_{G}$ and variance $\sigma_{G}$:
\begin{equation}
    \mu_{G} = \frac{1}{|E|} \sum_{e \in E} w, \
    \quad \sigma^{2}_{G} = \frac{1}{|E|}\sum_{e \in E} (w-\mu_{G})^{2}.
\end{equation}

Then, the mean distance and variance distance between the mobility graph $G_{a}$ and $G_{b}$ are expressed as:
\begin{align}
    D_{mean}(G_{a}, G_{b}) &= \left | \mu_{G_{a}} - \mu_{G_{b}} \right |, \\
    D_{var}(G_{a}, G_{b}) &= \left | \sigma^{2}_{G_{a}} - \sigma^{2}_{G_{b}} \right |.
\end{align}

It is not sufficient to only compare the mean and variance. For example, The mean and variance of the morning and evening peak hours are similar, but their destinations are different (working and residential areas, respectively). Specifically, to pay attention to flow imbalance of two regions, we first propose the unidirectional flow distance:
\begin{equation}\label{eq:flow_dist}
    D_{unif}(G_{a}, G_{b}) = \left | UniF(G_{a})-UniF(G_{a}) \right |,  
\end{equation}
where $UniF(G_{t}) = \sum_{v_i \in V} \sum_{v_j \in V} \left | \omega^t_{ij} - \omega^t_{ji} \right |$ is the unidirectional flow index of the graph, and $D_{unif}(G_{a}, G_{b}) = D_{unif}(G_{b}, G_{a})$.


Then, to highlight the high-weight edges in mobility graph $G_{t}$, we encode $G_{t}$ as a spatial structure label matrix $\mathcal{E}^{t}$, where each element $\mathcal{E}^{t}_{ij}$ represents whether the weight $\omega^{t}_{ij}$ of edge $e^{t}_{ij}$ in $G_{t}$ is large enough:
\begin{align}\label{Eq.4}
    \mathcal{E}^{t}_{ij} & = 
        \begin{cases}
         1 & {\omega^{t}_{ij}>\mu_{ij}}  \\
         0 & {\omega^{t}_{ij}\leq\mu_{ij}}
        \end{cases},
\end{align}
where $\mu_{ij} = \frac{1}{T}\sum_{t=1}^{T} \mu^{t}_{ij}$ represents the mean value of edge weight between node $v_i$ and $v_j$ over the time series.

Afterwards, the spatial structure encoding distance between mobility graphs $G_{a}$ and $G_{b}$ is defined as follows:
\begin{equation}
    D_{ss}(G_{a}, G_{b}) = ||\mathcal{E}_{a} \oplus \mathcal{E}_{b}||_{1},
\end{equation}
where $\oplus$ is the xor operation.

\subsubsection{Temporal Aggregation with Mobility Graph Distance}
After calculating the distances between mobility graphs, considering mobility graphs with close temporal distance are similar, we take temporal similarity into account. We define mobility graph distance (MGD) between mobility graphs $G_{a}$ and $G_{b}$ as the sum of above distance weighted by the temporal similarity with non-linearity, given as follows:
\begin{equation}
    MGD(G_{a}, G_{b}) = Z(\Delta t) \sum c_{i} \mathrm{M} (D_{i}),
\end{equation}
where $c_{i}$ is the weight of $i$-th distance, $D_{i}$ could be $D_{mean}, D_{var}, D_{unif}, D_{ss}$, $\mathrm{M}(\cdot)$ denotes the normalization function such as MinMaxScaler, $Z(\cdot)$ is an activation function. $\Delta t$ denotes the time interval between $G_{a}$ and $G_{b}$. 

Finally, we calculate the distance between mobility graphs through MGD, and use the clustering method (i.e., hierarchical cluster) to aggregate mobility graphs with different patterns thus to generate mobility patterns $\mathcal{G}$. 

\subsection{Mobility Pattern Joint Learning}
\begin{figure}[htbp]
	\centerline{\includegraphics[width=0.5\textwidth]{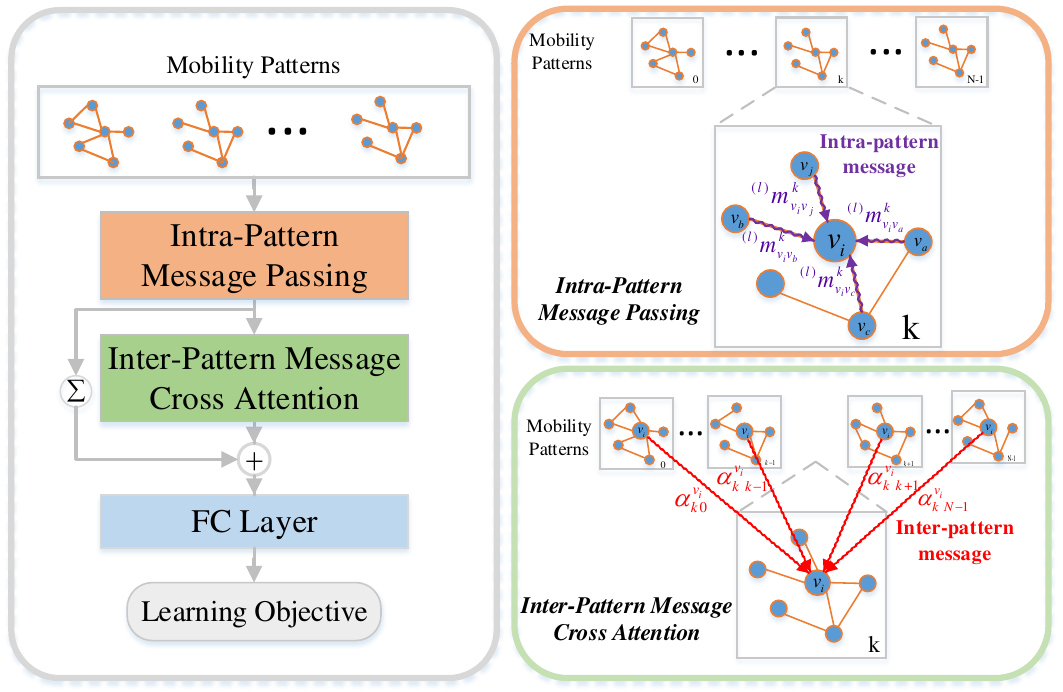}}
	\caption{The architecture of mobility pattern joint learning module (left) mainly consists: 1) intra-pattern massage passing (upper right), 2) inter-pattern message cross attention (down right). The former aggregates intra-pattern message to generate region embeddings in $N$ patterns. The latter obtains a fused region embeddings by attention mechanism using inter-pattern message.}
	\label{fig.3}
\end{figure}

In order to learn the underlying information of urban regions from mobility patterns, we present a mobility pattern joint learning module as shown in Figure \ref{fig.3}. The framework mainly consists of two parts: intra-pattern message passing and inter-pattern message cross attention. In the first part, the region hidden representations are updated by intra-pattern message propagating and aggregating in each mobility pattern thus extracting the spatial correlations between regions. In the second part, the cross-attention mechanism is used to integrate the inter-pattern message between the different mobility patterns for each region. Comprehensive embeddings output by two parts with residual connections \cite{resnet2016} are integrated by a fully connected layer to generate the final region embeddings.

\subsubsection{Intra-Pattern Message Passing}
The property of an urban region is affected by other regions with different impacts due to the spatial correlations. We assume that such impact is highly correlated to the human mobility condition of a region. To capture the spatial correlations, inspired by \cite{mpnn2017}, we design a intra-pattern message passing layer to propagate the inner flow messages between different regions in each mobility pattern. 

This part takes mobility patterns $\mathcal{G}$ as input. We initialize the $0^{th}$-layer hidden representations of regions $v_i$ in mobility pattern $\mathcal{G}_k$ as input region features $x^{k}_{v_i}$, i.e.: $^{(0)}h_{v_i}^k = x^{k}_{v_i}$.
Considering the mobility flow from region $v_j$ to region $v_i$, inspired by \cite{gat2017}, we compute the message sent along the mobility flow edge $v_j \rightarrow v_i$ by self-attention mechanism. The message function is defined as follows:
\begin{equation}
    m^k_{v_i v_j} = \frac{\exp \left( \left \langle W_{\mathrm{q}^p} h^{k}_{v_i}, W_{\mathrm{k}^p} h^{k}_{v_j} \right \rangle /\sqrt{d} \right)}{\textstyle \sum_{v_b \in N(v_i)}\exp \left ( \left \langle W_{\mathrm{q}^p} h^{k}_{v_i}, W_{\mathrm{k}^p} h^{k}_{v_b}  \right \rangle /\sqrt{d} \right )}, 
\end{equation}
where $\left \langle \cdot , \cdot \right \rangle$ is inner product, $ W_{\mathrm{q}^p}$ and $ W_{\mathrm{k}^p}$ are two trainable projection matrices, and $N(v_i)$ is neighbor region set of $v_i$. Here, the message from region $v_j$ to $v_i$ is also attention score.

Afterwards, region $v_i$ aggregates the messages sent by its neighbor regions $N(v_i)$, and updates its hidden representation by a weighed sum from $N(v_i)$, given as follows.
\begin{equation}
^{(l)}h_{v_i}^k  = {\textstyle \sum_{v_j \in N(v_i)}} {m^k_{v_i v_j}} ^{(l)}W_{\mathrm{v}^p} ^{(l-1)}h_{v_j}^k,  
\end{equation}
where $^{(l)}W_{\mathrm{v}^p}$ is a learnable projection matrix in $l^{th}$ layer. Here, as the mobility pattern is a directed complete graph, the neighbors of each region are all the regions.

To stabilize the learning process and learn information at different conceptual levels, we extend the self-attention mechanism to be multi-head ones \cite{attention2017}. Specifically, we concatenate $F$ parallel attention message functions with different learnable projections:
\begin{equation}
^{(l)}h_{v_i}^k = \left |  \right | _{f=1}^{F} \left \{ {\textstyle \sum_{v_j \in V}} {^{(f)}m^k_{v_i v_j}} {^{(f,\ l)}W_{\mathrm{v}^p}} ^{(l-1)}h_{v_j}^k \right \}, 
\end{equation}
where $\left | \right |$ represents concatenation operation, and${^{(f)}m^k_{v_i v_j}}$ represents the attention score calculated by the message function in the $f^{th}$ head attention. 

Taking the importance of mobility directions and the difference between in-flow and out-flow mobility for a region into account, we divide each mobility pattern $\mathcal{G}_k$ into two parts: source mobility pattern $\mathcal{G}_k^s$ and target mobility pattern $\mathcal{G}_k^t$. Here, we regard the out-flow and in-flow mobility as node features of source intra-pattern $X^s_k$ and target intra-pattern $X^t_k$, and apply intra-message passing layer in these two type of patterns, respectively, where $X^s_k = {X^t_k}^{\top}$. To fuse these two region representations, we project the concatenation of them to generate hidden representation of $v_i$:
\begin{equation}
    ^{(l)}h_{v_i}^k = f({^{(l)}_s}h_{v_i}^k \left | \right| {^{(l)}_t}h_{v_i}^k)
\end{equation}
where $f(\cdot)$ represents a linear projection. The region hidden representations $h_{v_i} = \{ h_{v_i}^0, h_{v_i}^1, ..., h_{v_i}^{N-1} \}$ extracting spatial dependency of $N$ mobility patterns are generated by stacking $L$ intra-pattern message passing layers.


\subsubsection{Inter-Pattern Message Cross Attention}
Intra-pattern message passing provides a robust mechanism for capturing the spatial correlations between regions in each mobility pattern. For learning the cross interactions between different mobility patterns for a region, we allow region to interact across mobility patterns via a self-attention mechanism \cite{attention2017}. 

Considering the region $v_i$ in mobility pattern $\mathcal{G}_a$ and $\mathcal{G}_b$, let the node features in $\mathcal{G}_a$ and $\mathcal{G}_b$ are $h_{v_i}^a$ and $h_{v_i}^b$, respectively. Then, for every such pair, we compute the correlation between $\mathcal{G}_a$ and $\mathcal{G}_b$ using attention as follows:
\begin{equation}
    \alpha_{ab}^{v_i} = \frac{\exp \left( \left \langle  W_{\mathrm{q}^c} h^{a}_{v_i},  W_{\mathrm{k}^c} h^{b}_{v_i} \right \rangle /\sqrt{d} \right)}{\textstyle \sum_{k=0}^{N-1} \exp \left ( \left \langle  W_{\mathrm{q}^c} h^{a}_{v_i},  W_{\mathrm{k}^c} h^{k}_{v_i}  \right \rangle /\sqrt{d} \right )}, 
\end{equation} 
where $W_{\mathrm{q}^c}$ and $W_{\mathrm{k}^c}$ are two trainable projection matrices, and the attention score $\alpha_{ab}^{v_i}$ reveals how $\mathcal{G}_a$ attends to the features of region $v_i$ of $\mathcal{G}_b$. Afterwards, we compute an inter-pattern message for the region $v_i$ of mobility pattern $\mathcal{G}_a$ by a weighed sum, where the multi-head attention is applied again:
\begin{equation}
    {h'}_{v_i}^a = \left |  \right | _{f=1}^{F} \left \{ {\textstyle \sum_{k=0}^{N-1}} {\alpha^{v_i}_{a k}}  W_{\mathrm{v}^c} h_{v_i}^k \right \}, 
\end{equation}
where $W_{\mathrm{v}^c}$ is a learnable projection matrix.

Finally, the fused region embedding of $v_i$ is updated by aggregating the inter-pattern message with mean aggregator:
\begin{equation}
    \bar{h}_{v_i} = \frac{1}{N} \sum_{k=0}^{N-1} {h'}_{v_i}^k.
\end{equation}

\subsubsection{Objective Function}
With the residual connection of intra-pattern message passing, we use a fully connected layer as an output layer to obtain the final region embeddings $\hat{h}_{v_i}$:
\begin{equation}
    \hat{h}_{v_i} = f(\frac{1}{N} \sum_{k=0}^{N-1} {h}_{v_i}^k + \bar{h}_{v_i}).
\end{equation}

Following ~\cite{HDGE2017}, we use region embeddings to estimate the distribution of mobility, and learn the embedding by minimizing the difference between the true distribution and the estimated distribution.
Given the source $v_{i}$, we calculate the transition probability of destination $v_{j}$:
\begin{equation}
    p_{\omega}(v_{j}|v_{i}) = \frac{\omega_{ij}}{\sum_{ v_{j^{*}} \in N(v_{i})} \omega_{ij^{*}}}.
\end{equation}
 
Then, given the region embedding $\hat{h}_{v_i}$, $\hat{h}_{v_j}$ for region $v_{i}$, $v_{j}$, we estimate the transition probability: 
\begin{equation}
    \hat{p}_{\omega}(v_{j}|v_{i}) = \frac{\exp({\hat{h}_{v_i}}^{\top} \hat{h}_{v_j})} {\sum_{ v_{j^{*}} \in N(v_{i})} \exp({\hat{h}_{v_i}}^{\top} \hat{h}_{v{_j^{*}}})}, 
\end{equation}

Finally, the objective function can expressed as:
\begin{equation}
    \mathcal{L} = \sum_{i,j} -p_{\omega}(v_{j}|v_{i})\log\hat{p}_{\omega}(v_{j}|v_{i}).
\end{equation}

\section{Evaluation}
This section aims to answer the following research questions:
\paragraph{RQ1} How is the performance of out MGFN as compared to various state-of-the-art methods?
\paragraph{RQ2} How do different components (e.g., mobility graph fusion, mobility pattern joint learning) affect the results?
\paragraph{RQ3} Can the proposed mobility graph fusion module really discover mobility patterns? What are its advantages compared with other methods?
\paragraph{RQ4} Why did other models perform worse than ours? Is it true that the other methods learned how traffic flow changes? (as we assume in Section ~\ref{sec.1})
\subsection{Experiment Settings}
\paragraph{Data Description}
We evaluate the performance of our method on New York City (NYC) datasets from NYC open data website \footnote{opendata.cityofnewyork.us}. \textbf{Census blocks} gives the boundaries of 180 regions split by streets in Manhattan, NYC. \textbf{Taxi trips} describes around 10 million taxi trip records during one month in the studied area. \textbf{Crime count} consists of around 40 thousand crime records during one year in the studied area. \textbf{Check-in count} contains over 100 thousand check-in locations of over 200 fine-grained categories. \textbf{Land usage type} divides the borough of Manhattan into 12 districts by the community boards. We follow the settings in ~\cite{MVURE2020} and apply taxi trip data as human mobility data and take the crime count, check-in count, land usage type as prediction tasks, respectively.
\paragraph{Baselines}
We compare MGFN with the following baseline methods: (1) \textbf{node2vec} ~\cite{node2vec_2016} uses biased random walks to learn node latent representations by skip-gram models; (2) \textbf{LINE} ~\cite{LINE2015} optimizes the objective function that preserves both the local and global network structures.
(3) \textbf{HDGE} ~\cite{HDGE2017} jointly embeds a spatial graph and a flow graph with temporal dynamics.
(4) \textbf{ZE-Mob} ~\cite{ZE-Mob2018} captures massive human mobility patterns, and models spatio-temporal co-occurrence of zones in the embedding learning; (5) \textbf{MV-PN} ~\cite{MV-PN2019} learns region embeddings with multi-view PoI network within the region; (6) \textbf{MVURE} ~\cite{MVURE2020} enable cross-view information sharing and weighted multi-view fusion with human mobility and inherent region attributes data (e.g. POI, check-in).

\paragraph{Parameter Settings} Following ~\cite{MVURE2020}, the dimension of region embeddings $d$ is 96. In mobility graph fusion module, the weight $c_i$ in MGD is set as 1, and the number of mobility patterns $N$ is set as 7. In mobility pattern joint learning module, the number of layers $L$ is set as 1. 

\subsection{Performance Comparison (RQ1)}

\begin{table*}[!t]
	\renewcommand{\arraystretch}{1.3}
	\caption{Performance comparison with different methods 
	for crime prediction, land usage classification and check-in count prediction tasks.}
	\label{tab.1}
	\centering
	\scalebox{0.88}{
	\begin{tabular}{ccccccccc}
		\toprule
		& \multicolumn{3}{c}{Crime Prediction} & \multicolumn{2}{c}{Land Usage Classification} & \multicolumn{3}{c}{Check-in Prediction} \\
		\cline{2-9}
		& MAE & RMSE & R$^2$ & NMI & ARI & MAE & RMSE & R$^2$\\
		\midrule
		LINE & 117.53 & 152.43 & 0.06 & 0.17 & 0.01 & 564.59 & 853.82 & 0.08 \\
		node2vec & 75.09 & 104.97 & 0.49 & 0.58 & 0.35 & 372.83 & 609.47 & 0.44 \\ 
		HDGE & 72.65 & 96.36  & 0.58 & 0.59 & 0.29 & 399.28 & 536.27 & 0.57 \\ 
		ZE-Mob & 101.98 & 132.16 & 0.20 & 0.61 & 0.39 & 360.71 & 592.92 & 0.47 \\ 
		MV-PN & 92.30 & 123.96 & 0.30 & 0.38 & 0.16 & 476.14 & 784.25 & 0.08 \\ 
		MVURE & 69.28 $\pm$ 3.13 & 96.51 $\pm$ 4.73 & 0.57 $\pm$ 0.04 & \textbf{0.78 $\pm$ 0.02} & \textbf{0.62 $\pm$ 0.06} & 312.63 $\pm$ 9.61 & 513.02 $\pm$ 16.64 & 0.61 $\pm$ 0.03 \\
		MGFN & 70.21 $\pm$ 2.31 & \textbf{89.60 $\pm$ 2.50} & \textbf{0.63 $\pm$ 0.02} & 0.75 $\pm$ 0.01 & 0.57 $\pm$ 0.01 & \textbf{292.60 $\pm$ 17.05} & \textbf{451.76 $\pm$ 28.12} & \textbf{0.69 $\pm$ 0.04} \\
		\bottomrule
	\end{tabular}
	}
\end{table*}
For regression tasks (i.e., crime, check-in), we apply the Lasso regression~\cite{LassoRegression} with metrics of Mean Absolute Error (MAE), Root Mean Square Error (RMSE) and coefficient of determination (R$^2$). For the clustering task (i.e., land usage classification), we use K-means to cluster region embeddings with Normalized Mutual Information (NMI) and Adjusted Rand Index (ARI) with settings in~\cite{ZE-Mob2018}.
Table \ref{tab.1} shows the results of crime prediction, land usage classification and check-in count prediction. We observe that: (1) urban region embedding approaches outperform traditional graph embedding methods, and graph modeling methods (i.e., MG-FN, MVURE) generally perform better than HDGE, ZE-Mob and MV-PN, indicating the spatial dependency between regions is a necessity for urban region embedding; and (2) our MGFN with only human mobility data outperforms the second best model MVURE with multi-sourced data (POI, check-in, mobility, etc.) and achieves up to 13.11\% improvement in terms of $R^{2}$ in the check-in prediction task. It is noted that multi-sourced data used in MVURE is always unreachable or accompanied by many noise data in real urban application.

\subsection{Model Ablation Study (RQ2)}
\begin{figure}[!t]
\centerline{\includegraphics[width=0.5\textwidth]{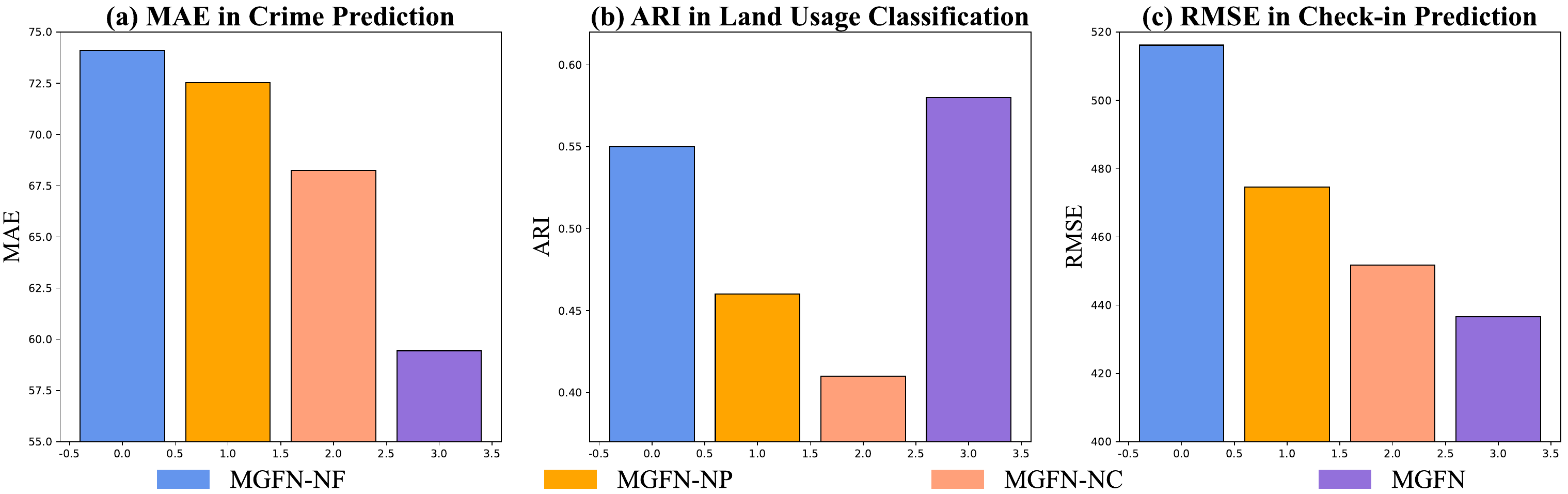}}
\caption{Ablation studies for three tasks on NYC dataset. (a) MAE in Crime Prediction. (b) ARI in Land Usage Classification. (c) RMSE in Check-in Prediction. }
\label{fig.4}
\end{figure}
To further investigate the effect of each component in our model, we compare MGFN with its three variants by removing multi graph fusion module, intra-pattern message passing and inter-pattern message cross attention in mobility pattern joint learning module from our method, which are named as MGFN-NF, MGFN-NP and MGFN-NC, respectively. Figure~\ref{fig.4} shows the MAE, ARI and RMSE results of MGFN and its variants in crime prediction, land usage classification and check-in prediction tasks respectively. We observe that MGFN performs better than MGFN-NF, demonstrating that the MGF module effectively eases the redundant temporal information for learning more general representations of urban regions. Moreover, MGFN consistently outperforms MGFN-NP and MGFN-NC, which indicates the effectiveness of intra message passing and aggregating between nodes in each mobility pattern and inter messages fusion between mobility patterns in modeling the complex spatial correlations.

\subsection{Graph Similarity Measurement (RQ3)}

\begin{figure}[!t]
	\centerline{\includegraphics[width=0.5\textwidth]{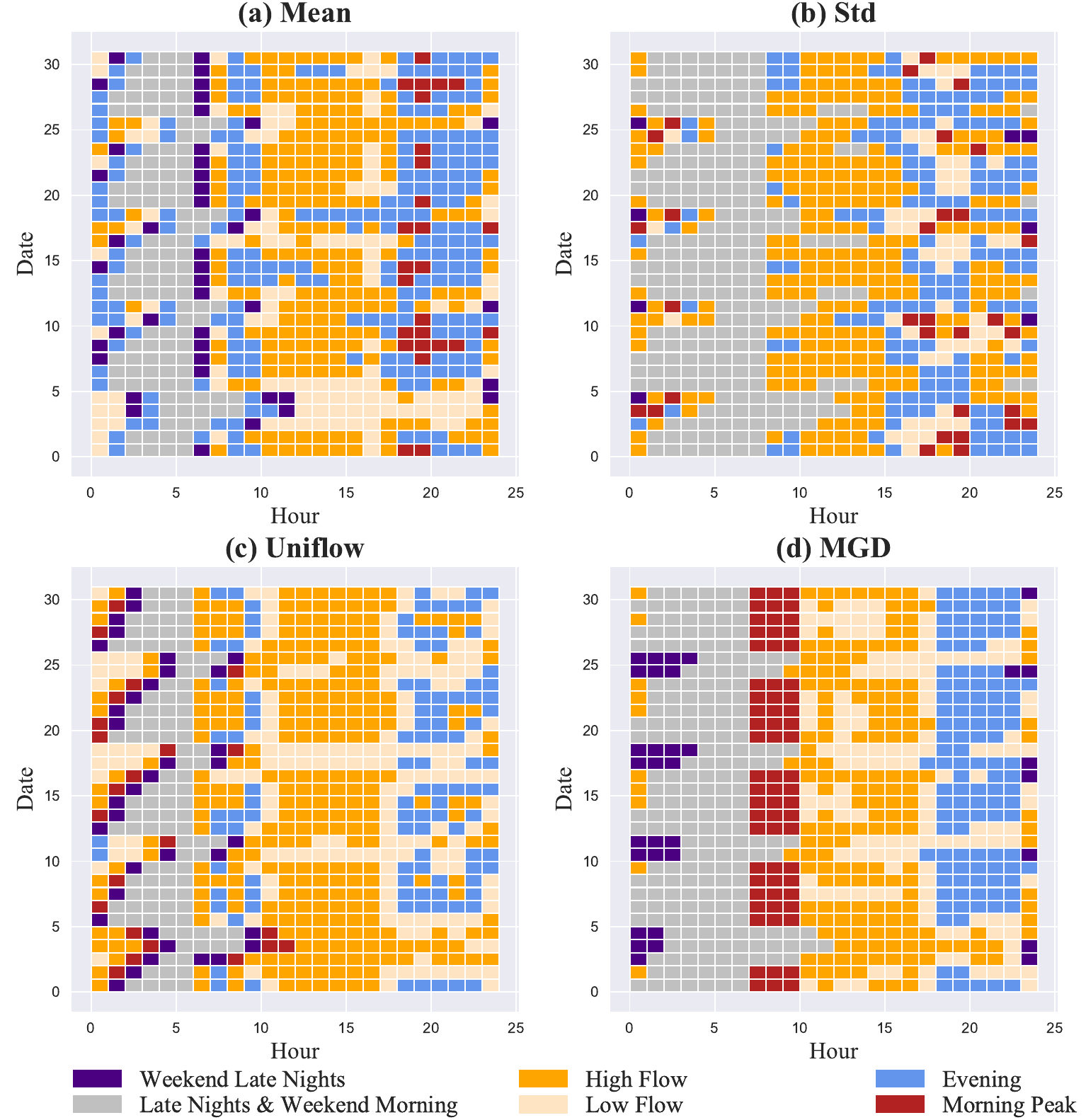}}
	\caption{Comparison of clustering results using different distances. The x-axis represents hour of a day, and the y-axis is the day of a month. Each sub-figure represents the result of a measurement, respectively. Each color represents a certain pattern corresponding to the law of human mobility. \textbf{Compared with other measurements, only MGD could effectively identify important patterns} such as morning peaks (red block, from 7 to 9 am for week-days), weekends late night (purple block from 12 pm to 2 am for week-ends), etc.}
	\label{fig.5}
\end{figure} 
%
To intuitively evaluate the performance of the proposed Mobility Graph Fusion module (Section~\ref{sec.3.1}), we visualize the mobility patterns extracted by our multi-graph distance (MGD) measurement compared with three measures in Section~\ref{sec.3.1} including Mean, Std, and Uniflow using hierarchical clustering. Specifically, in Figure~\ref{fig.5}, a certain color represents a specific pattern corresponding to the law of human mobility. We observe that different from other measures, our MGD (Figure 5.d) is able to distinguish patterns of both morning/evening peaks and working days/weekends effectively, due to our integrated considerations with spatio-temporal information. We argue that this improvement is beneficial to enabling the cross-domain prediction (i.e., from traffic prediction to crime prediction) by uncovering the dynamic correlations contained in urban regional functionality.

\subsection{Generalization Ability of MGFN (RQ4)} 
\begin{figure}[!t]
	\centerline{\includegraphics[width=0.5\textwidth]{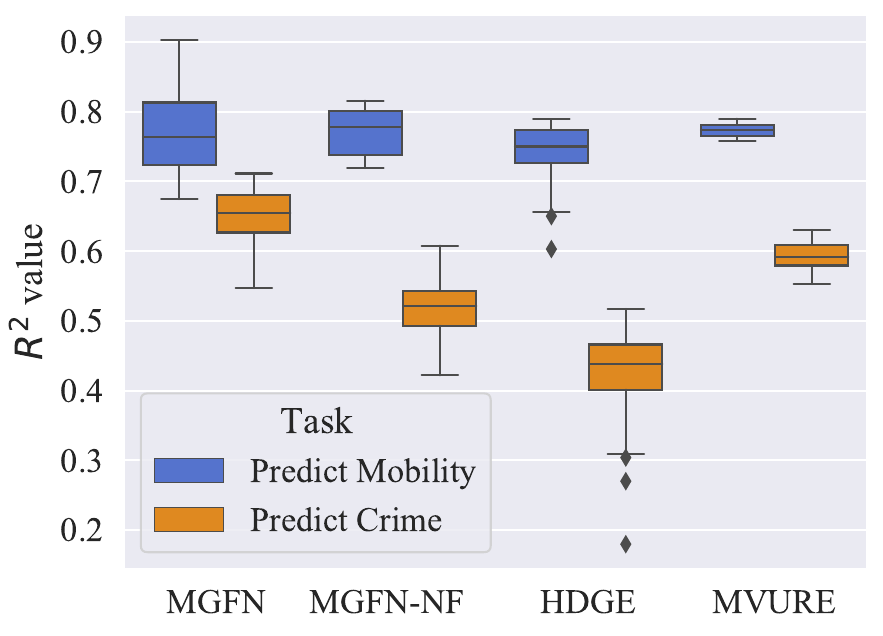}}
	\caption{The generalization ability of our MGFN compared with three other methods. 
	}
	\label{fig.6}
\end{figure}

To show the generalization ability of our method, we compare our MGFN with other three methods on both supervised task (i.e., mobility prediction) and cross-domain task (i.e., crime prediction) in Figure~\ref{fig.6}. Other methods perform poorly on cross-domain tasks because they learn how traffic changes rather than a generalized region representation. 
Moreover, after removed MGF, the generalization performance is reduced by about 20$\%$, which shows the importance of MGF.



\section{Related Works}
\paragraph{Graph Representation Learning} Graph embedding aims to learn a low-dimensional vector by mapping the characteristics of nodes to a low-dimensional vector space so that the proximities of nodes can be well preserved ~\cite{survey_ne_2018}. Early works devote to learn the shallow representations by graph factorization approaches ~\cite{le_2001} relying on spectral embedding from graph Laplacian and skip-gram based methods ~\cite{deepwalk_2014,node2vec_2016,LINE2015} learned by random walk objectives. More recently, graph neural networks (GNNs) have become a widely used tool for graph embedding ~\cite{gat2017,mpnn2017}. 

\paragraph{Region Representation Learning} Several strategies have been studied to learn the representation of regions. The first strategy learns embeddings from time-series human mobility data. HDGE~\cite{HDGE2017} uses fine-grained human mobility to construct flow graph, and learn the region embedding at different times. ZE-mob~\cite{ZE-Mob2018} regards the region as a word and the mobility event as context, and learn the embedding via a word embedding method. The second strategy learns embeddings from multi-view cross-geo-type(region and other spatio-temporal items) correlations. Fu \textit{et al.}~\shortcite{MV-PN2019} take into account both intra-region structural information and inter-region spatial auto-correlations. Zhang \textit{et al.} ~\shortcite{MVURE2020} use a cross-view information sharing method to learn comprehensive region embeddings. 

\section{Conclusion}
In this paper, we focus on learning generalized embeddings to enable cross domain urban computing tasks. We proposed: (1) a novel mobility graph fusion method where redundant graphs are integrated as patterns; and (2) a novel mobility pattern joint learning method to enable the cross graph embeddings that mutually enhance each other and provide more effective representation for downstream tasks. Extensive experiments on real-world mobility data show the proposed MGFN outperforms all baseline methods. Besides, in-depth analysis reveals insightful observations, e.g., Late nights and evenings on weekends show a different pattern from workdays. In future, we will extend our framework to other downstream tasks (e.g., house price prediction). 


\section*{Acknowledgments}
The research is supported by Natural Science Foundation of China (61872306), and Fundamental Research Funds for the Central Universities (20720200031).

\bibliographystyle{named}
\bibliography{ijcai22}

\end{document}